%% file: main.tex
\documentclass[runningheads]{llncs}
\usepackage[T1]{fontenc}
\usepackage{graphicx}
\usepackage{subcaption}
\usepackage{amsmath}
\usepackage{multirow}
\begin{document}
\title{Structured Definitions and Segmentations for Legal Reasoning in LLMs: A Study on Indian Legal Data}

%
%
\author{Mann Khatri\inst{1}\orcidID{0000-0002-5132-9223} \and
 Mirza Yusuf\inst{1}\orcidID{0000-0002-8293-5381} \and
Rajiv Ratn Shah\inst{1}\orcidID{0000-0003-1028-9373} \and
Ponnurangam Kumaraguru\inst{2}\orcidID{0000-0001-5082-2078}
}
\authorrunning{Mann et al.}
%
\institute{Indraprastha Institute of Information Technology, Delhi\\
\email{\{mannk,rajivratn\}@iiitd.ac.in, mirzayusuf1000@gmail.com} \and
International Institute of Information Technology Hyderabad\\
\email{pk.guru@iiit.ac.in}\\
}
\maketitle              
\begin{abstract}
Large Language Models (LLMs), trained on extensive datasets from the web, exhibit remarkable general reasoning skills. Despite this, they often struggle in specialized areas like law, mainly because they lack domain-specific pretraining. The legal field presents unique challenges, as legal documents are generally long and intricate, making it hard for models to process the full text efficiently. Previous studies have examined in-context approaches to address the knowledge gap, boosting model performance in new domains without full domain alignment. In our paper, we analyze model behavior on legal tasks by conducting experiments in three areas: (i) reorganizing documents based on rhetorical roles to assess how structured information affects long context processing and model decisions, (ii) defining rhetorical roles to familiarize the model with legal terminology, and (iii) emulating the step-by-step reasoning of courts regarding rhetorical roles to enhance model reasoning. These experiments are conducted in a zero-shot setting across three Indian legal judgment prediction datasets. Our results reveal that organizing data or explaining key legal terms significantly boosts model performance, with a minimum increase of ~1.5\% and a maximum improvement of 4.36\% in F1 score compared to the baseline.

\end{abstract}
\section{Introduction}

Large Language Models (LLMs) have exhibited impressive generalization capabilities across a broad spectrum of natural language processing tasks \cite{brown2020language,zhang2023instruction}. Their ability to follow instructions, reason over complex inputs, and generate coherent text has made them powerful tools for downstream applications in diverse domains \cite{choi2018quac,rajpurkar2016squad,dua2019drop}. This success is largely attributed to pretraining on massive and heterogeneous datasets, which enables LLMs to capture a wide range of linguistic and semantic patterns.

However, the computational and financial costs associated with pretraining such models are substantial. As a result, domain-specific LLMs are relatively rare, and most applications rely on general-purpose models that may not perform optimally in specialized contexts. This limitation is particularly evident in fields such as biomedicine, law, and finance, where domain knowledge and terminology diverge significantly from the data used during pretraining \cite{zhao2022charge,mukherjee2023orca}.

To mitigate this domain mismatch without incurring the high cost of retraining, recent studies have explored in-context learning (ICL) as a strategy to guide LLMs at inference time \cite{mu-etal-2024-navigating}. By embedding relevant in-context information like examples into the prompt, ICL enables models to adapt to new tasks with minimal supervision. These approaches have shown encouraging results across a range of general tasks, demonstrating that prompt engineering can, to some extent, compensate for the lack of domain alignment.

The legal domain, and in particular legal judgment prediction (LJP), poses unique challenges for LLMs. Legal documents are often lengthy, formally structured, and replete with specialized vocabulary and rhetorical conventions \cite{kalamkar-etal-2022-corpus}, . Predicting outcomes based on such documents requires not only linguistic competence but also an understanding of domain-specific reasoning, procedural logic, and hierarchical structure \cite{zhao2022charge,zhang2023case,wu-etal-2023-precedent,wei2022chain,mukherjee2023orca}. These complexities make legal tasks especially demanding in zero-shot or few-shot settings \cite{jiang2023legal}.

While there has been growing interest in applying ICL to the legal domain, prior work has primarily focused on leveraging factual context or case retrieval \cite{mu-etal-2024-navigating}. Some studies have explored the use of exemplars or retrieved precedents to improve performance on legal reasoning tasks \cite{wu2022towards}.

The extent to which restructuring or explicitly clarifying legal concepts can improve LLMs' abilities in zero-shot LJP tasks has been inadequately explored. This research investigates how prompt design, especially by integrating legal rhetorical roles and structured reasoning, can enhance LLMs' effectiveness in legal judgment predictions. We introduce three methodological approaches: (i) reorganizing documents based on rhetorical roles to help the model navigate through complex legal texts more easily; (ii) defining rhetorical roles to familiarize the model with specialized vocabulary; and (iii) mimicking court-like reasoning to replicate the logical flow of legal arguments. These methods are assessed in a zero-shot context using three datasets related to Indian legal judgment predictions. Our results demonstrate that some of these minimal modifications improve performance, highlighting the importance of using structurally and semantically enriched prompts for LLMs in specialized fields. Additionally, the results shed light on the impact of each component both separately and in combination. Ultimately, it was found that including all components did not yield the best outcomes, while shorter combinations proved more effective.

\section{Related Work}


\subsection{Zero-Shot, Few-Shot, CoT} Advanced techniques like zero-shot, few-shot, and chain-of-thought (CoT) prompting have been utilized to enhance LJP model performance. Zero-shot learning allows models to predict unseen cases without prior similar examples, leveraging vast pre-existing knowledge \cite{brown2020languagemodelsfewshotlearners}. Few-shot learning improves predictions by exposing models to a few examples during training, bridging the gap between zero-shot and fully supervised learning \cite{gao2021makingpretrainedlanguagemodels}. CoT prompting refines prediction by encouraging models to generate intermediate reasoning steps, mimicking human thought processes \cite{wei2022chain}. These methodologies show promise in improving the accuracy and interpretability of legal judgment predictions.

Recent studies applied these techniques to legal reasoning tasks like the Japanese Bar exam through the COLIEE entailment task \cite{rabelo2022overview}. CoT prompting and fine-tuning with explanation approaches improved performance, with the most significant improvements using prompts derived from legal reasoning techniques like IRAC (Issue, Rule, Application, Conclusion). Zero-shot and few-shot learning with legal reasoning prompts demonstrated promising but variable results, necessitating further investigation and careful selection of few-shot examples \cite{yu2022legalpromptingteachinglanguage}.

\subsection{Legal Judgement Prediction}
Early efforts in LJP have employed neural networks to forecast outcomes in English courts, revealing challenges in language understanding and contextual reasoning within legal texts \cite{chalkidis-etal-2019-neural}. Building on this, further work extended neural approaches to the UK legal system, offering insights into how unique aspects of British jurisprudence influence judgment outcomes \cite{10.1145/3388176.3388183}.

Moving beyond monolingual legal systems, research in the Swiss context has emphasized the importance of multilingual and multicultural considerations. By providing a benchmark dataset for LJP across multiple languages and legal traditions, this work underscores the necessity of localized resources for building robust and generalizable models \cite{niklaus-etal-2021-swiss}.

To improve model interpretability and accuracy, several studies have focused on integrating domain-specific legal knowledge into neural architectures. For instance, legal knowledge graphs and structured representations have been used to enrich case understanding and improve predictive performance \cite{Gan_Kuang_Yang_Wu_2021}. This direction is further refined through multi-stage representation learning techniques designed to capture the hierarchical and nuanced structure of legal documents \cite{10.1145/3404835.3462945}.

In the works of \cite{nigam2024rethinking}, this line of inquiry is extended to the Indian legal domain under a realistic setting, where predictions are made using only the information available at the time of decision—such as facts, statutes, precedents, and arguments. Transformer-based models (InLegalBERT, BERT, XLNet) and LLMs (LLaMA-2, GPT-3.5 Turbo) are systematically evaluated, with the latter showing strong performance. The study highlights the importance of incorporating legal context and introduces human evaluation metrics—Clarity and Linking—to assess the quality of predictions and explanations. While results are promising, the study concludes that current LLMs still fall short of expert-level performance.


\cite{malik-etal-2021-ildc} introduces a corpus tailored to the Indian judicial system, underscoring the importance of culturally and linguistically specific datasets in improving prediction models' relevance and accuracy. Innovative models like \cite{nigam2024legaljudgmentreimaginedpredex} and \cite{Almuslim2024CanAL} demonstrate specialized AI systems' potential to interpret and predict legal judgments in specific national contexts, showcasing LJP research's evolving landscape and practical applications.

\subsection{LLM Reasoning} Studies have examined LLMs' reasoning capabilities, highlighting advancements and challenges. \cite{qiao-etal-2023-reasoning} provides a comprehensive overview of LLM reasoning approaches, categorizing them into strategic and knowledge enhancements. This survey describes reasoning strategies like CoT prompting and advanced techniques integrating human-like reasoning with external computation engines to boost performance. \cite{huang-chang-2023-towards} analyzes current methodologies, challenges, and benchmarks development to measure reasoning abilities, discussing future directions to bridge the gap between LLM capabilities and human-like reasoning.


\subsection{Legal LLM Reasoning} Studies have highlighted the potential and challenges in exploring LLMs' reasoning capabilities within a legal context. \cite{Almeida_2024} analyzes LLMs' capacity for moral and legal reasoning across eight experiments, comparing their responses to human participants. While LLMs often align with human decision-making factors, significant variation in correlations indicates that LLM psychology diverges from human psychology systematically.

\cite{nay2023largelanguagemodelstax} demonstrates LLMs' potential to autonomously identify relevant legal authorities and apply them to specific scenarios, particularly in US tax law. While current models do not yet match professional lawyers' expertise, they could significantly assist in legal work, increasing productivity and reducing costs. These findings suggest that LLMs could eventually perform a wide range of legal tasks more accurately, prompting potential regulatory changes in legal services and AI governance.

\section{Background}

Legal case proceedings follow a structured sequence, beginning with the plaintiff's complaint, the defendant's response, and pretrial activities such as discovery and motions. During the trial, both parties present evidence and examine witnesses, followed by closing arguments. The judge or jury then gives a verdict that can be appealed. In delivering judgments, judges analyze the facts of the case, previous rulings, evidence, and legal arguments to apply the relevant law and articulate the rationale behind their decisions.

However, judgment documents often lack a clear structure, making it difficult to trace the reasoning process \cite{kalamkar-etal-2022-corpus}. To address this, we use \textit{Rhetorical Roles}, which assign semantic functions to sentences, such as facts, arguments, precedents, and decisions, to uncover the underlying logic. Our ablations are built upon this structure, mirroring the judge’s reasoning process for better document understanding.

Following this, our prompt consists of three main components:
\begin{itemize}
    \item \textbf{Rhetorical Roles (R):} Sentences assigned to specific rhetorical roles are combined to form a paragraph within the prompt. There will be N paragraphs corresponding to the N rhetorical roles identified in the judgment document, concatenated during model input. Note that not all roles may appear in a judgment document. Each paragraph is preceded by a heading that corresponds to its role. The final document format resembles the following: \textit{[FAC]\textbackslash{n}\{sentences from the FAC role\}\textbackslash{n}\textbackslash{n}[RLC]\textbackslash{n}\{sentences from the RLC role...\}}
    \item \textbf{Definition (D):} This component appears at the start of the prompt, offering definitions of rhetorical roles to assist the LLM in understanding the legal jargon.
    \item \textbf{Chain (C):} Upon receiving input, the LLM generates an ANALYSIS. The ANALYSIS and the initial input are then fed back into the LLM to produce the RATIO. Afterwards, the original input, ANALYSIS, and RATIO are all input into the LLM to generate the RPC for the case. This recursive approach mimics court processes. This method of reasoning through chaining is referred to as chain throughout the paper.
\end{itemize}

\section{Formulation and Data Preparation}

\subsection{Dataset}

In our research, we utilize the dataset of two rhetorical roles compiled by \cite{kalamkar-etal-2022-corpus} and \cite{bambroo2025marro}, which provides an organized assortment of cases from Indian judiciary records. This primary dataset, containing 50 samples, classifies judgment documents into 13 rhetorical categories (comprising 12 main categories plus a NONE category), whereas the secondary dataset, containing 20 samples, highlights 7 crucial rhetorical roles vital for logical reasoning. In our experiment, we exclude the rhetorical roles ANALYSIS, STA, RATIO, and RPC as inputs to the model on the assumption that the model will generate these roles to arrive at the verdict.

To further assess the efficiency of our legal judgment prediction approach we chose a bigger dataset, Predex dataset \cite{nigam2024legaljudgmentreimaginedpredex}, which comprises approximately 12,000 test samples. This dataset presents input cases by omitting analysis and outcomes, enabling the model to anticipate these elements for the LJP. It should be noted that this dataset does not include rhetorical role annotation, confining us to experiment with definitions and chains.

\begin{figure*}[!ht]
    \centering
    \includegraphics[width=\linewidth]{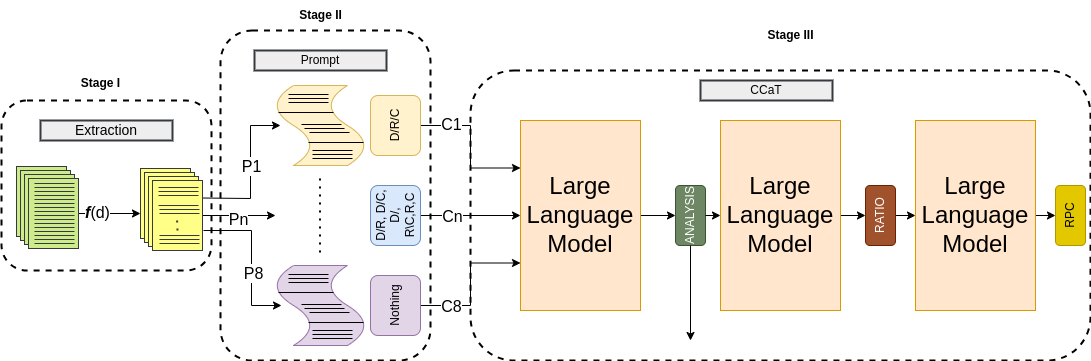}
    \caption{The process consists of three stages. In the first stage, documents are segregated into Rhetorical Roles (P1, .., P8), which are then incorporated into the prompt in the second stage. Depending on the prompt requirements, either segregated or unsegregated cases are used. In the third stage, these prompts (C1, .., C8) are fed into the LLM one to three times based on the need for chaining. The final explanation of the model is noted after this stage.}
    \label{fig:working}
\end{figure*}

\subsection{Annotation}
In our study, two annotators manually annotated 70 (50+20) legal cases present in the test set of first and all from the second dataset, assigning binary labels (0 or 1) based on the outcome of each case, as the judgments were clearly indicated at the end of each document. Given the explicit nature of the verdicts, the involvement of a legal expert was deemed unnecessary. To maintain consistency and avoid ambiguity, we excluded cases with partially appealed judgments, removing six cases from the dataset. Before finalizing the ground-truth labels, disagreements between the authors regarding the annotations were resolved through discussion. Finally, we were left with 64 (44+20) cases in the test set.

The rationale for choosing only the test set for zero-shot inference is to ensure that if the dataset is employed in the future for training or few-shot inference, the training set can be used accordingly, thus avoiding the need to label new documents.

\subsection{Task Formulation}
Given a set of documents $D=\{d_1, d_2, d_3, ..., d_n\}$, a function $f$ that modifies the sentences of the document in their corresponding rhetorical roles paragraphs, a function $g$ that adds the definitions of the rhetorical roles in the prompt. We also add the chaining component to obtain the reasoning as shown in figure \ref{fig:working}. We perform ablations on these components to see the effect of every component in the legal understading.
\\
\begin{align}
    y_{\text{CASE with chain}} &= \mathbf{w}\left( \text{RPC} \mid \mathbf{v}\left( \text{RATIO} \mid \mathbf{u}\left( \text{ANALYSIS} \mid P_i \right) \right) \right) \\
    y_{\text{CASE without chain}} &= \mathbf{u}\left( \text{ANALYSIS} \mid P_i \right)
\end{align}

Here, $\textbf{u}(.)$, $\textbf{v}(.)$, $\textbf{w}(.)$ are the LLMs and $P_i$ are the corresponding prompts. The output of the LLM, $y\in\{YES, NO\}$, is recorded in each step. The value taken by $y$ is $YES$ if the case in favour of the plaintiff, else $NO$.

In an effort to shorten the prompt by few tokens, we initially eliminated the PREAMBLE rhetorical role and assessed the ensuing effects and its significance on the decision-making process. The absence of PREAMBLE negatively affected the model's performance, likely because the model relies on meta-data such as the competing parties' names typically found in the PREAMBLE. Consequently, we decided to retain PREAMBLE in our experiments. This shows that model finds it difficult to spot plaintiffs and defendants.

\input{Both_dataset_table}
\section{Experiments}
For our experimentation, we chose the instruct models of different sizes Mistral 7B (Mistral-7B-Instruct-v0.3) \cite{jiang2023mistral}, Phi3 (Phi-3-mini-128k-instruct) \cite{abdin2024phi}, Llama-3.1 (Llama-3.1-8B-instruct) \cite{touvron2023llama} and o3-mini\footnote{\url{https://openai.com/index/openai-o3-mini/}} (o3-mini-2025-01-31) as they have large context windows. The prompt with the most tokens was from \textit{with definition with roles with chain (D/R/C)} variant, with more than 10k tokens, as it has the most information of all variants mentioned. Each model was tested using a zero-shot approach, as detailed in the following ablations. It is important to note that, after obtaining explanations from each ablation, we subsequently prompted the model to determine if the case decision favored the plaintiff or not for a binary output. This additional step was implemented to ensure consistency between the explanation and the output, preventing discrepancies such as an analysis indicating a win for the plaintiff while the output suggests otherwise. Such errors were observed sporadically when this extra step was omitted. Note that in the case of non-chained ablation, the $YES$ or $NO$ is output by the model based only on the ANALYSIS generated. 

To ensure the reproducibility of the results, we set \textit{do\_sample} to \textit{False}. After analyzing the section lengths in the ground truth of the training set cases, we determined that the value for the hyperparameter \textit{max\_new\_tokens} should be 2000. For o3-mini since it lacked a configuration for making the outputs deterministic so we ran the model 5 times and reported the results.

\input{Both_dataset_table_expl}
\begin{table}[!ht]
\centering
\begin{tabular}{c|c|c|c|c}
\textbf{Metrics}  & \textbf{D / C} & \textbf{D}      & \textbf{C}      & \textbf{None}   \\ \hline
\textbf{Macro F1} & 0.6689         & 0.6447          & \textbf{0.6732} & 0.6296                                                                \\
\textbf{FPR}      & 0.4576         & 0.5495          & \textbf{0.4151} & 0.4314                                                                     \\
\textbf{FNR}      & 0.1401         & \textbf{0.095}  & 0.1695          & 0.2394                                                                   
\end{tabular}
\caption{Prediction results on Predex dataset. D means presence of definitions and C means whether chains were used. D/C is the combination of both and None is none of the methods were included in the prompt.}
\label{predex}
\end{table}

\subsection{Evaluation metrics}
For the task of judgment prediction, we chose macro-F1, false negative rate (FNR) and false positive rate (FPR) as metrics. FNR and FPR is to ensure ``presumption of innocence" that systems do not find an innocent person guilty and hence we try to minimize FNR and FPR while maximizing F1.

For explaination metrics, we opted for ROUGE-1, ROUGE-2 \cite{lin2004rouge}, METEOR \cite{banerjee2005meteor} and BertScore \cite{zhang2019bertscore}. Using both ROUGE and METEOR provides comprehensive evaluation due to their complementary strengths. ROUGE excels at recall-oriented tasks like summarization, capturing reference content overlap, while METEOR considers both precision and recall, penalizing word order mistakes for a human-like judgment. METEOR also addresses nuances like synonyms and paraphrases, offering a more balanced and nuanced assessment. This dual approach ensures better alignment with human evaluation metrics. Whereas, on the other hand BertScore is used to capture the semantic meaning of the summaries.

\section{Results}

\subsection{Judgment Prediction}
For Dataset1, the most effective ablation configuration for LLaMA involved the use of rhetorical role definitions alongside their segmentation (D/R), without chaining prompts. This configuration was closely followed by one that employed definitions and chaining (D/C) but excluded segmentation. Notably, the combination of definitions and segmentation (D/R) resulted in the lowest false positive rate (FPR) and false negative rate (FNR). A similar trend was observed in Dataset2, where the optimal setup also incorporated both definitions and segmentation (D/R). While the lowest FPR (0) was achieved in this configuration, it coincided with a relatively high FNR of 62.5. Conversely, the lowest FNR (12.5) was obtained in the definitions-only (D) ablation, albeit with a significantly higher FPR. A balanced trade-off between FPR and FNR was achieved in the configuration that included both definitions and segmentation of rhetorical roles (D/R). See Table \ref{both_datasets}

In Dataset1, the Phi-3 model yielded results close to random baselines and performed poorly in terms of macro-F1 score. See Phi in Table \ref{both_datasets}. Among its variants, the configuration incorporating chaining (C) performed best, though all FPRs exceeded 40 and FNRs ranged between 25 and 43. While this represented a marginal improvement over the \textit{None} ablation, Phi-3’s overall performance remained suboptimal. For Dataset2, Phi-3 performed below random, with the best-performing variant combining definitions and chaining (D/C), achieving an F1 score of 34.84. This configuration also exhibited a better balance between FPR and FNR relative to other ablations.

In Dataset1, models were expected to output binary decisions (i.e., \texttt{YES} or \texttt{NO}); however, Mistral-3 occasionally generated indecisive responses. In contrast, Phi-3, LLaMA-3.1, and o3-mini consistently produced binary outputs. The indecisiveness observed in Mistral-3 (Row 1 of Mistral results) likely reflects the model’s uncertainty in cases lacking sufficient evidence. The introduction of chaining reduced the frequency of such undecided predictions. To address inconsistencies, we filtered the evaluation set to include only those examples for which all chained variants produced decisive outputs, resulting in a subset of 15 examples (Row 2 in Mistral). Given the small sample size, we performed an additional ablation by comparing each configuration to its non-chained counterpart, focusing exclusively on decisive examples. Across all Mistral ablations, the configuration that integrated both definitions and segmentation (D/R) consistently outperformed others—across overall, common, and chain-wise filtering—yielding the lowest FPR and FNR values. For Dataset2, the best-performing variant was the one using only rhetorical role definitions (D), which also achieved the lowest FPR and FNR.

Due to the stochastic nature of OpenAI’s model APIs, each ablation was executed five times, and outputs were aggregated. In Dataset1, the best-performing ablation in terms of consistency and performance was the configuration involving definitions and chaining (D/C). It also exhibited lower variance compared to other configurations. The lowest FPR was achieved by the ablation combining definitions with segmentation (D/R), whereas the lowest FNR was observed in the configuration with definitions and chaining (D/C). The next best overall configuration included both definitions and segmentation of rhetorical roles (D/R).

For Dataset3 (i.e., the PredEx dataset), only the LLaMA model was evaluated due to the subpar performance of Phi and Mistral on previous datasets, as well as cost constraints associated with OpenAI models. Among the LLaMA variants, the chained (C) configuration demonstrated strong performance, though it was closely matched by the definitions and chained (D/C) variant in terms of F1 score. The FPR and FNR values for both variants were also comparable, with only minor trade-offs observed between the two. See Table \ref{predex}.

This suggests that, in general, definitions of rhetorical roles significantly contributed to the model's enhanced performance by augmenting its understanding of the complex structure of legal language within both datasets. Llama demonstrated robust performance across all predictive tasks, whereas Phi exhibited the least success. Notably, the ablation involving all components did not yield the best performance outcomes.

\subsection{Judgment Explanation}
The ROUGE scores were nearly identical across all models, with Llama-3.1 slightly outperforming Mistral and o3-mini in Dataset1, which in turn outperformed Phi. See Table \ref{both_datasets_expl}. o3-mini performed slightly better on Rouge in Dataset2 when compared to Llama and Mistral. However, these scores did not fully capture the results, necessitating a qualitative analysis. For qualitative analysis (QA) We randomly sampled cases that were common True Positives (TP) and True Negatives (TN) across all models. While Llama-3.1, Phi and o3-mini shared False Positive (FP) and False Negative (FN) cases, Mistral had no overlapping FP and FN cases, so we randomly sampled these for Mistral. In total, we analyzed six cases.

In QA, a True Positive (TP) case for Llama-3.1 and o3-mini, the predicted analysis closely matched the ground truth. In the example, the judge deduced that the State Government had set a higher rate for stamp duty on the premises, which the model failed to infer. Otherwise, the predicted analysis was consistent with the ground truth. This observation was also applicable to the Mistral and Phi models.

For the TN case, key points were identified across all models. Generally, Llama-3.1 functioned more like a judge, while Mistral and Phi focused on analyzing the provided information in general manner rather than analysing legally. Among other two, o3-mini and Mistral outperformed Phi, though Llama-3.1 demonstrated a greater overlap in judgment.

In true cases, we observed that there was a good overlap between the explanation provided by the model and the ground truth but the wordings and placement of the sentences hardly matched between the two.

The QA for FP case, for Llama-3.1 and Phi, the outputs were more general rather than legal in their analysis. Phi's output was slightly better than Llama-3.1's, as Phi mentioned potential courses of action the court could take in its decision-making process. In both cases, the ground analysis had minimal overlap with the predicted analysis. For Mistral, there was no overlap between the ground truth and the analysis. Moreover, it relied on older precedents to reach a conclusion, as the severity of the crime was not able to outweigh the precedents.

For FN cases of Llama-3.1 and Phi in QA, the models made mistakes while considering precedents and seemed unaware of the underlying law. Both models erred in their explanations, providing almost identical reasons for the decision. For Mistral, it was observed the same as in the FP case, there was no overlap between the ground truth and the analysis. Moreover, it relied on older precedents to reach a conclusion, as the severity of the crime was not able to outweigh the precedents.

The findings suggest that Llama-3.1 and o3-mini aligns more closely with legal analysis, whereas Mistral and Phi tend to provide more generic analyses that are not as closely tied to the legal context. This indicates that Llama-3.1 follows instructions better compared to the other two models, as the system prompt clearly specifies the role of the LLM.

\section{Conclusion}
This study investigated the influence of defining legal jargon, segmentation, and the reasoning processes applied by the judicial method. The analysis was performed on three datasets utilizing models of different scales, highlighting the significance of employing False Positive Rate (FPR) and False Negative Rate (FNR) for model evaluation, especially considering the legal doctrine of the "presumption of innocence." Additionally, the research illustrates the capability of models to interpret legal language in a zero-shot context, as demonstrated with Llama-3.1 and o3-mini, which adhered rigorously to legal standards.

\section{Limitations and Future Work}
The two evaluation dataset was relatively small due to limited data availability and the requirement for manual evaluation. The limited memory capacity of our GPUs restricted us from processing longer sentences in the language model. This limitation prevented us from using more complex and extended input sequences, such as incorporating other legal documents like precedents and related laws. This also restricted our ability to fine-tune or pre-train these models.

Future work can explore few-shot and fine-tuning methods to enhance the model's understanding of legal terminology. If we obtain a larger dataset, we can examine the temporal aspects of the legal domain. Additionally, instead of only considering binary outcomes, we can explore partial appeals to make systems more useful and robust.

%
%
%
\bibliographystyle{splncs04}
\bibliography{biblio}
%




\end{document}

%% file: Both_dataset_table.tex
\begin{table}[!htbp]
\resizebox{\textwidth}{!}{
\begin{tabular}{|c|ccccccccc|}
\hline
\multirow{2}{*}{\textbf{Models}}  & \multicolumn{1}{l|}{}             & \multicolumn{1}{c|}{\textbf{D/R/C}}                                                                 & \multicolumn{1}{c|}{\textbf{D/R}}                                                                            & \multicolumn{1}{c|}{\textbf{D/C}}                                                                   & \multicolumn{1}{c|}{\textbf{D}}                                                                              & \multicolumn{1}{c|}{\textbf{R/C}}                                                                   & \multicolumn{1}{c|}{\textbf{R}}                                                                     & \multicolumn{1}{c|}{\textbf{C}}                                                                    & \textbf{None}                                                                  \\ \cline{2-10} 
                                  & \multicolumn{9}{c|}{\textbf{Dataset 1}}                                                                                                                                                                                                                                                                                                                                                                                                                                                                                                                                                                                                                                                                                                                                                                                                                                       \\ \hline
\multirow{3}{*}{\textbf{Llama}}   & \multicolumn{1}{c|}{\textbf{FPR}} & \multicolumn{1}{c|}{36.84}                                                                          & \multicolumn{1}{c|}{\textbf{28}}                                                                             & \multicolumn{1}{c|}{28.57}                                                                          & \multicolumn{1}{c|}{28.57}                                                                                   & \multicolumn{1}{c|}{30.77}                                                                          & \multicolumn{1}{c|}{36.84}                                                                          & \multicolumn{1}{c|}{33.33}                                                                         & 33.33                                                                          \\ \cline{2-10} 
                                  & \multicolumn{1}{c|}{\textbf{FNR}} & \multicolumn{1}{c|}{40}                                                                             & \multicolumn{1}{c|}{\textbf{21.05}}                                                                          & \multicolumn{1}{c|}{30.43}                                                                          & \multicolumn{1}{c|}{40}                                                                                      & \multicolumn{1}{c|}{41.94}                                                                          & \multicolumn{1}{c|}{40}                                                                             & \multicolumn{1}{c|}{41.38}                                                                         & 30                                                                             \\ \cline{2-10} 
                                  & \multicolumn{1}{c|}{\textbf{f1}}  & \multicolumn{1}{c|}{61.36}                                                                          & \multicolumn{1}{c|}{\textbf{75}}                                                                             & \multicolumn{1}{c|}{70.45}                                                                          & \multicolumn{1}{c|}{63.64}                                                                                   & \multicolumn{1}{c|}{61.36}                                                                          & \multicolumn{1}{c|}{61.36}                                                                          & \multicolumn{1}{c|}{61.36}                                                                         & 68.18                                                                          \\ \hline
\multirow{3}{*}{\textbf{Phi}}     & \multicolumn{1}{c|}{\textbf{FPR}} & \multicolumn{1}{c|}{48.65}                                                                          & \multicolumn{1}{c|}{46.88}                                                                                   & \multicolumn{1}{c|}{44.44}                                                                          & \multicolumn{1}{c|}{47.37}                                                                                   & \multicolumn{1}{c|}{47.5}                                                                           & \multicolumn{1}{c|}{47.5}                                                                           & \multicolumn{1}{c|}{\textbf{40}}                                                                   & 43.33                                                                          \\ \cline{2-10} 
                                  & \multicolumn{1}{c|}{\textbf{FNR}} & \multicolumn{1}{c|}{42.86}                                                                          & \multicolumn{1}{c|}{41.67}                                                                                   & \multicolumn{1}{c|}{\textbf{25}}                                                                    & \multicolumn{1}{c|}{33.33}                                                                                   & \multicolumn{1}{c|}{\textbf{25}}                                                                    & \multicolumn{1}{c|}{\textbf{25}}                                                                    & \multicolumn{1}{c|}{28.57}                                                                         & 35.71                                                                          \\ \cline{2-10} 
                                  & \multicolumn{1}{c|}{\textbf{f1}}  & \multicolumn{1}{c|}{46}                                                                             & \multicolumn{1}{c|}{52.07}                                                                                   & \multicolumn{1}{c|}{54.48}                                                                          & \multicolumn{1}{c|}{47.62}                                                                                   & \multicolumn{1}{c|}{45.41}                                                                          & \multicolumn{1}{c|}{45.41}                                                                          & \multicolumn{1}{c|}{\textbf{62.39}}                                                                & 57.69                                                                          \\ \hline
\multirow{3}{*}{\textbf{Mistral}} & \multicolumn{1}{c|}{\textbf{FPR}} & \multicolumn{1}{c|}{\textbf{\begin{tabular}[c]{@{}c@{}}10.0/\\ 0.0/\\ 0.0\end{tabular}}}            & \multicolumn{1}{c|}{\textbf{\begin{tabular}[c]{@{}c@{}}10.0/\\ 0.0/\\ 0.0\end{tabular}}}                     & \multicolumn{1}{c|}{\begin{tabular}[c]{@{}c@{}}35.29/\\ 20.0/\\ 35.29\end{tabular}}                 & \multicolumn{1}{c|}{\begin{tabular}[c]{@{}c@{}}31.58/\\ 14.29/\\ 31.58\end{tabular}}                         & \multicolumn{1}{c|}{\begin{tabular}[c]{@{}c@{}}26.67/\\ 0.0/\\ 28.57\end{tabular}}                  & \multicolumn{1}{c|}{\begin{tabular}[c]{@{}c@{}}41.18/\\ 16.67/\\ 41.18\end{tabular}}                & \multicolumn{1}{c|}{\begin{tabular}[c]{@{}c@{}}33.33/\\ 0.0/\\ 35.0\end{tabular}}                  & \begin{tabular}[c]{@{}c@{}}39.13/\\ 0.0/\\ 38.1\end{tabular}                   \\ \cline{2-10} 
                                  & \multicolumn{1}{c|}{\textbf{FNR}} & \multicolumn{1}{c|}{\begin{tabular}[c]{@{}c@{}}20.0/\\ 20.0/\\ 33.33\end{tabular}}                  & \multicolumn{1}{c|}{\textbf{\begin{tabular}[c]{@{}c@{}}0.0/\\ 0.0/\\ 11.11\end{tabular}}}                    & \multicolumn{1}{c|}{\begin{tabular}[c]{@{}c@{}}30.0/\\ 30.0/\\ 45.0\end{tabular}}                   & \multicolumn{1}{c|}{\begin{tabular}[c]{@{}c@{}}12.5/\\ 12.5/\\ 38.89\end{tabular}}                           & \multicolumn{1}{c|}{\begin{tabular}[c]{@{}c@{}}20.0/\\ 20.0/\\ 29.41\end{tabular}}                  & \multicolumn{1}{c|}{\begin{tabular}[c]{@{}c@{}}22.22/\\ 22.22/\\ 35.71\end{tabular}}                & \multicolumn{1}{c|}{\begin{tabular}[c]{@{}c@{}}20.0/\\ 20.0/\\ 35.0\end{tabular}}                  & \begin{tabular}[c]{@{}c@{}}20.0/\\ 20.0/\\ 36.84\end{tabular}                  \\ \cline{2-10} 
                                  & \multicolumn{1}{c|}{\textbf{f1}}  & \multicolumn{1}{c|}{\begin{tabular}[c]{@{}c@{}}63.18 (36)/\\ 86.11 (15)/\\ 77.5 (18)\end{tabular}}  & \multicolumn{1}{c|}{\textbf{\begin{tabular}[c]{@{}c@{}}89.44 (19)/\\ 100.0 (15)/\\ 94.43 (18)\end{tabular}}} & \multicolumn{1}{c|}{\begin{tabular}[c]{@{}c@{}}58.95 (39)/\\ 72.22 (15)/\\ 59.46 (37)\end{tabular}} & \multicolumn{1}{c|}{\begin{tabular}[c]{@{}c@{}}68.22 (41)/\\ 86.61 (15)/\\ 64.76 (37)\end{tabular}}          & \multicolumn{1}{c|}{\begin{tabular}[c]{@{}c@{}}65.02 (44)/\\ 86.11 (15)/\\ 70.85 (31)\end{tabular}} & \multicolumn{1}{c|}{\begin{tabular}[c]{@{}c@{}}61.25 (31)/\\ 79.64 (15)/\\ 61.25 (31)\end{tabular}} & \multicolumn{1}{c|}{\begin{tabular}[c]{@{}c@{}}66.67 (42)/\\ 86.11 (15)/\\ 65.0 (40)\end{tabular}} & \begin{tabular}[c]{@{}c@{}}61.82 (42)/\\ 86.11 (15)/\\ 62.48 (40)\end{tabular} \\ \hline
\multirow{3}{*}{\textbf{o3}}      & \multicolumn{1}{c|}{\textbf{FPR}} & \multicolumn{1}{c|}{\begin{tabular}[c]{@{}c@{}}30.0\\ ±6.1\end{tabular}}                            & \multicolumn{1}{c|}{\textbf{\begin{tabular}[c]{@{}c@{}}24.55\\ ±7.61\end{tabular}}}                          & \multicolumn{1}{c|}{\begin{tabular}[c]{@{}c@{}}28.18\\ ±3.8\end{tabular}}                           & \multicolumn{1}{c|}{\begin{tabular}[c]{@{}c@{}}27.27\\ ±3.21\end{tabular}}                                   & \multicolumn{1}{c|}{\begin{tabular}[c]{@{}c@{}}30.91\\ ±5.93\end{tabular}}                          & \multicolumn{1}{c|}{\textbf{\begin{tabular}[c]{@{}c@{}}24.55\\ ±4.07\end{tabular}}}                 & \multicolumn{1}{c|}{\begin{tabular}[c]{@{}c@{}}31.82\\ ±3.21\end{tabular}}                         & \begin{tabular}[c]{@{}c@{}}28.18\\ ±5.93\end{tabular}                          \\ \cline{2-10} 
                                  & \multicolumn{1}{c|}{\textbf{FNR}} & \multicolumn{1}{c|}{\begin{tabular}[c]{@{}c@{}}52.73\\ ±6.89\end{tabular}}                          & \multicolumn{1}{c|}{\begin{tabular}[c]{@{}c@{}}52.73\\ ±6.89\end{tabular}}                                   & \multicolumn{1}{c|}{\textbf{\begin{tabular}[c]{@{}c@{}}40.0\\ ±7.47\end{tabular}}}                  & \multicolumn{1}{c|}{\begin{tabular}[c]{@{}c@{}}45.45\\ ±7.87\end{tabular}}                                   & \multicolumn{1}{c|}{\begin{tabular}[c]{@{}c@{}}50.0\\ ±5.57\end{tabular}}                           & \multicolumn{1}{c|}{\begin{tabular}[c]{@{}c@{}}54.55\\ ±5.57\end{tabular}}                          & \multicolumn{1}{c|}{\begin{tabular}[c]{@{}c@{}}50.91\\ ±7.47\end{tabular}}                         & \begin{tabular}[c]{@{}c@{}}50.91\\ ±4.98\end{tabular}                          \\ \cline{2-10} 
                                  & \multicolumn{1}{c|}{\textbf{f1}}  & \multicolumn{1}{c|}{\begin{tabular}[c]{@{}c@{}}58.01\\ ±4.86\end{tabular}}                          & \multicolumn{1}{c|}{\begin{tabular}[c]{@{}c@{}}60.54\\ ±6.58\end{tabular}}                                   & \multicolumn{1}{c|}{\textbf{\begin{tabular}[c]{@{}c@{}}65.71\\ ±3.38\end{tabular}}}                 & \multicolumn{1}{c|}{\begin{tabular}[c]{@{}c@{}}63.27\\ ±4.7\end{tabular}}                                    & \multicolumn{1}{c|}{\begin{tabular}[c]{@{}c@{}}59.16\\ ±5.46\end{tabular}}                          & \multicolumn{1}{c|}{\begin{tabular}[c]{@{}c@{}}59.51\\ ±4.3\end{tabular}}                           & \multicolumn{1}{c|}{\begin{tabular}[c]{@{}c@{}}58.21\\ ±5.16\end{tabular}}                         & \begin{tabular}[c]{@{}c@{}}59.87\\ ±3.08\end{tabular}                          \\ \hline
\multicolumn{1}{|l|}{}            & \multicolumn{9}{c|}{\textbf{Dataset 2}}                                                                                                                                                                                                                                                                                                                                                                                                                                                                                                                                                                                                                                                                                                                                                                                                                                       \\ \hline
\multirow{3}{*}{\textbf{Llama}}   & \multicolumn{1}{c|}{\textbf{FPR}} & \multicolumn{1}{c|}{25}                                                                             & \multicolumn{1}{c|}{25}                                                                                      & \multicolumn{1}{c|}{25}                                                                             & \multicolumn{1}{c|}{75}                                                                                      & \multicolumn{1}{c|}{50}                                                                             & \multicolumn{1}{c|}{25}                                                                             & \multicolumn{1}{c|}{\textbf{0}}                                                                    & \textbf{0}                                                                     \\ \cline{2-10} 
                                  & \multicolumn{1}{c|}{\textbf{FNR}} & \multicolumn{1}{c|}{37.5}                                                                           & \multicolumn{1}{c|}{31.25}                                                                                   & \multicolumn{1}{c|}{50}                                                                             & \multicolumn{1}{c|}{\textbf{12.5}}                                                                           & \multicolumn{1}{c|}{37.5}                                                                           & \multicolumn{1}{c|}{37.5}                                                                           & \multicolumn{1}{c|}{50}                                                                            & 62.5                                                                           \\ \cline{2-10} 
                                  & \multicolumn{1}{c|}{\textbf{f1}}  & \multicolumn{1}{c|}{60.11}                                                                          & \multicolumn{1}{c|}{\textbf{64.29}}                                                                          & \multicolumn{1}{c|}{52}                                                                             & \multicolumn{1}{c|}{56.71}                                                                                   & \multicolumn{1}{c|}{52.38}                                                                          & \multicolumn{1}{c|}{60.11}                                                                          & \multicolumn{1}{c|}{58.33}                                                                         & 49.49                                                                          \\ \hline
\multirow{3}{*}{\textbf{Phi}}     & \multicolumn{1}{c|}{\textbf{FPR}} & \multicolumn{1}{c|}{25}                                                                             & \multicolumn{1}{c|}{25}                                                                                      & \multicolumn{1}{c|}{25}                                                                             & \multicolumn{1}{c|}{\textbf{0}}                                                                              & \multicolumn{1}{c|}{25}                                                                             & \multicolumn{1}{c|}{\textbf{0}}                                                                     & \multicolumn{1}{c|}{25}                                                                            & \textbf{0}                                                                     \\ \cline{2-10} 
                                  & \multicolumn{1}{c|}{\textbf{FNR}} & \multicolumn{1}{c|}{87.5}                                                                           & \multicolumn{1}{c|}{100}                                                                                     & \multicolumn{1}{c|}{\textbf{75}}                                                                    & \multicolumn{1}{c|}{93.75}                                                                                   & \multicolumn{1}{c|}{100}                                                                            & \multicolumn{1}{c|}{93.75}                                                                          & \multicolumn{1}{c|}{87.5}                                                                          & 87.5                                                                           \\ \cline{2-10} 
                                  & \multicolumn{1}{c|}{\textbf{f1}}  & \multicolumn{1}{c|}{24.81}                                                                          & \multicolumn{1}{c|}{13.04}                                                                                   & \multicolumn{1}{c|}{\textbf{34.84}}                                                                 & \multicolumn{1}{c|}{23.27}                                                                                   & \multicolumn{1}{c|}{13.04}                                                                          & \multicolumn{1}{c|}{23.27}                                                                          & \multicolumn{1}{c|}{24.81}                                                                         & 29.29                                                                          \\ \hline
\multirow{3}{*}{\textbf{Mistral}} & \multicolumn{1}{c|}{\textbf{FPR}} & \multicolumn{1}{c|}{\begin{tabular}[c]{@{}c@{}}75.0/\\ 50.0/\\ 66.67\end{tabular}}                  & \multicolumn{1}{c|}{\begin{tabular}[c]{@{}c@{}}62.5/\\ 50.0/\\ 62.5\end{tabular}}                            & \multicolumn{1}{c|}{\begin{tabular}[c]{@{}c@{}}75.0/\\ 60.0/\\ 71.43\end{tabular}}                  & \multicolumn{1}{c|}{\textbf{\begin{tabular}[c]{@{}c@{}}50.0/\\ 33.33/\\ 50.0\end{tabular}}}                  & \multicolumn{1}{c|}{\begin{tabular}[c]{@{}c@{}}70.0/\\ 60.0/\\ 70.0\end{tabular}}                   & \multicolumn{1}{c|}{\begin{tabular}[c]{@{}c@{}}66.67/\\ 60.0/\\ 66.67\end{tabular}}                 & \multicolumn{1}{c|}{\begin{tabular}[c]{@{}c@{}}66.67/\\ 50.0/\\ 66.67\end{tabular}}                & \begin{tabular}[c]{@{}c@{}}81.82/\\ 71.43/\\ 81.82\end{tabular}                \\ \cline{2-10} 
                                  & \multicolumn{1}{c|}{\textbf{FNR}} & \multicolumn{1}{c|}{\begin{tabular}[c]{@{}c@{}}11.11/\\ 11.11/\\ 16.67\end{tabular}}                & \multicolumn{1}{c|}{\begin{tabular}[c]{@{}c@{}}0.0/\\ 0.0/\\ 10.0\end{tabular}}                              & \multicolumn{1}{c|}{\begin{tabular}[c]{@{}c@{}}12.5/\\ 12.5/\\ 20.0\end{tabular}}                   & \multicolumn{1}{c|}{\begin{tabular}[c]{@{}c@{}}10.0/\\ 10.0/\\ 9.09\end{tabular}}                            & \multicolumn{1}{c|}{\begin{tabular}[c]{@{}c@{}}12.5/\\ 12.5/\\ 11.11\end{tabular}}                  & \multicolumn{1}{c|}{\begin{tabular}[c]{@{}c@{}}12.5/\\ 12.5/\\ 15.38\end{tabular}}                  & \multicolumn{1}{c|}{\textbf{\begin{tabular}[c]{@{}c@{}}0.0/\\ 0.0/\\ 0.0\end{tabular}}}            & \begin{tabular}[c]{@{}c@{}}16.67/\\ 16.67/\\ 14.29\end{tabular}                \\ \cline{2-10} 
                                  & \multicolumn{1}{c|}{\textbf{f1}}  & \multicolumn{1}{c|}{\begin{tabular}[c]{@{}c@{}}52.38 (20)/\\ 70.68 (13)/\\ 58.46 (18)\end{tabular}} & \multicolumn{1}{c|}{\begin{tabular}[c]{@{}c@{}}62.5 (18)/\\ 74.51 (13)/\\ 62.5 (18)\end{tabular}}            & \multicolumn{1}{c|}{\begin{tabular}[c]{@{}c@{}}51.28 (19)/\\ 63.89 (13)/\\ 52.96 (17)\end{tabular}} & \multicolumn{1}{c|}{\textbf{\begin{tabular}[c]{@{}c@{}}72.31 (18)/\\ 78.33 (13)/\\ 71.67 (17)\end{tabular}}} & \multicolumn{1}{c|}{\begin{tabular}[c]{@{}c@{}}54.76 (19)/\\ 63.89 (13)/\\ 54.76 (19)\end{tabular}} & \multicolumn{1}{c|}{\begin{tabular}[c]{@{}c@{}}59.29 (19)/\\ 63.89 (13)/\\ 59.29 (19)\end{tabular}} & \multicolumn{1}{c|}{\begin{tabular}[c]{@{}c@{}}62.5 (18)/\\ 74.51 (13)/\\ 62.5 (18)\end{tabular}}  & \begin{tabular}[c]{@{}c@{}}41.33 (20)/\\ 51.25 (13)/\\ 41.56 (18)\end{tabular} \\ \hline
\multirow{3}{*}{\textbf{o3}}      & \multicolumn{1}{c|}{\textbf{FPR}} & \multicolumn{1}{c|}{\begin{tabular}[c]{@{}c@{}}25.0\\ ±17.68\end{tabular}}                          & \multicolumn{1}{c|}{\begin{tabular}[c]{@{}c@{}}10.0\\ ±13.69\end{tabular}}                                   & \multicolumn{1}{c|}{\begin{tabular}[c]{@{}c@{}}25.0\\ ±0.0\end{tabular}}                            & \multicolumn{1}{c|}{\begin{tabular}[c]{@{}c@{}}25.0\\ ±0.0\end{tabular}}                                     & \multicolumn{1}{c|}{\begin{tabular}[c]{@{}c@{}}45.0\\ ±11.18\end{tabular}}                          & \multicolumn{1}{c|}{\begin{tabular}[c]{@{}c@{}}35.0\\ ±13.69\end{tabular}}                          & \multicolumn{1}{c|}{\begin{tabular}[c]{@{}c@{}}20.0\\ ±11.18\end{tabular}}                         & \textbf{\begin{tabular}[c]{@{}c@{}}0.0\\ ±0.0\end{tabular}}                    \\ \cline{2-10} 
                                  & \multicolumn{1}{c|}{\textbf{FNR}} & \multicolumn{1}{c|}{\begin{tabular}[c]{@{}c@{}}56.25\\ ±7.65\end{tabular}}                          & \multicolumn{1}{c|}{\begin{tabular}[c]{@{}c@{}}57.5\\ ±2.8\end{tabular}}                                     & \multicolumn{1}{c|}{\begin{tabular}[c]{@{}c@{}}60.0\\ ±8.39\end{tabular}}                           & \multicolumn{1}{c|}{\textbf{\begin{tabular}[c]{@{}c@{}}38.75\\ ±2.8\end{tabular}}}                           & \multicolumn{1}{c|}{\begin{tabular}[c]{@{}c@{}}51.25\\ ±5.23\end{tabular}}                          & \multicolumn{1}{c|}{\begin{tabular}[c]{@{}c@{}}58.75\\ ±3.42\end{tabular}}                          & \multicolumn{1}{c|}{\textbf{\begin{tabular}[c]{@{}c@{}}55.0\\ ±5.23\end{tabular}}}                 & \begin{tabular}[c]{@{}c@{}}52.5\\ ±5.59\end{tabular}                           \\ \cline{2-10} 
                                  & \multicolumn{1}{c|}{\textbf{f1}}  & \multicolumn{1}{c|}{\begin{tabular}[c]{@{}c@{}}47.76\\ ±6.72\end{tabular}}                          & \multicolumn{1}{c|}{\begin{tabular}[c]{@{}c@{}}50.65\\ ±3.09\end{tabular}}                                   & \multicolumn{1}{c|}{\begin{tabular}[c]{@{}c@{}}45.32\\ ±5.66\end{tabular}}                          & \multicolumn{1}{c|}{\textbf{\begin{tabular}[c]{@{}c@{}}59.3\\ ±1.82\end{tabular}}}                           & \multicolumn{1}{c|}{\begin{tabular}[c]{@{}c@{}}45.76\\ ±6.04\end{tabular}}                          & \multicolumn{1}{c|}{\begin{tabular}[c]{@{}c@{}}43.68\\ ±4.46\end{tabular}}                          & \multicolumn{1}{c|}{\begin{tabular}[c]{@{}c@{}}49.98\\ ±5.51\end{tabular}}                         & \begin{tabular}[c]{@{}c@{}}56.58\\ ±3.89\end{tabular}                          \\ \hline
\end{tabular}
}
\caption{Judgment Precition Results on Llama, Phi and o3-mini. F1, FPR and FNR represent Macro averaged F1, False Negative Rates and False Positive Rates respectively. In the case of the Mistral row, each cell contains three additional rows showing results for: (i) independent, where chains were evaluated with its corresponding binary outputs; (ii) common, involving unified test points across all chains that produced a binary result; and (iii) chainwise, which compares the performance of chained vs. non-chained  counterparts.}
\label{both_datasets}
\end{table}

%% file: Both_dataset_table_expl.tex
\begin{table}[!htbp]
\resizebox{\textwidth}{!}{
\begin{tabular}{|c|ccccccccc|}
\hline
\multirow{2}{*}{\textbf{Models}}  & \multicolumn{1}{l|}{}                   & \multicolumn{1}{c|}{\textbf{D/R/C}}                                                               & \multicolumn{1}{c|}{\textbf{D/R}}                                                                 & \multicolumn{1}{c|}{\textbf{D/C}}                                                                          & \multicolumn{1}{c|}{\textbf{D}}                                                                   & \multicolumn{1}{c|}{\textbf{R/C}}                                                                          & \multicolumn{1}{c|}{\textbf{R}}                                                                   & \multicolumn{1}{c|}{\textbf{C}}                                                                   & \textbf{None}                                                                \\ \cline{2-10} 
                                  & \multicolumn{9}{c|}{\textbf{Dataset 1}}                                                                                                                                                                                                                                                                                                                                                                                                                                                                                                                                                                                                                                                                                                                                                                                                                              \\ \hline
\multirow{3}{*}{\textbf{Llama}}   & \multicolumn{1}{c|}{\textbf{R1/R2}}     & \multicolumn{1}{c|}{\begin{tabular}[c]{@{}c@{}}40.3/\\ 16.44\end{tabular}}                        & \multicolumn{1}{c|}{\begin{tabular}[c]{@{}c@{}}38.0/\\ 15.63\end{tabular}}                        & \multicolumn{1}{c|}{\begin{tabular}[c]{@{}c@{}}40.5/\\ 16.83\end{tabular}}                                 & \multicolumn{1}{c|}{\begin{tabular}[c]{@{}c@{}}39.1/\\ 16.17\end{tabular}}                        & \multicolumn{1}{c|}{\begin{tabular}[c]{@{}c@{}}39.7/\\ 16.15\end{tabular}}                                 & \multicolumn{1}{c|}{\begin{tabular}[c]{@{}c@{}}41.4/\\ 16.99\end{tabular}}                        & \multicolumn{1}{c|}{\begin{tabular}[c]{@{}c@{}}40.6/\\ 16.46\end{tabular}}                        & \textbf{\begin{tabular}[c]{@{}c@{}}41.7/\\ 17.14\end{tabular}}               \\ \cline{2-10} 
                                  & \multicolumn{1}{c|}{\textbf{METEOR}}    & \multicolumn{1}{c|}{29.74}                                                                        & \multicolumn{1}{c|}{24.1}                                                                         & \multicolumn{1}{c|}{29.05}                                                                                 & \multicolumn{1}{c|}{22.8}                                                                         & \multicolumn{1}{c|}{\textbf{31.7}}                                                                         & \multicolumn{1}{c|}{29.8}                                                                         & \multicolumn{1}{c|}{30.94}                                                                        & 28.27                                                                        \\ \cline{2-10} 
                                  & \multicolumn{1}{c|}{\textbf{BertScore}} & \multicolumn{1}{c|}{83.38}                                                                        & \multicolumn{1}{c|}{\textbf{83.39}}                                                               & \multicolumn{1}{c|}{83.36}                                                                                 & \multicolumn{1}{c|}{83.37}                                                                        & \multicolumn{1}{c|}{83.21}                                                                                 & \multicolumn{1}{c|}{83.21}                                                                        & \multicolumn{1}{c|}{83.45}                                                                        & 83.45                                                                        \\ \hline
\multirow{3}{*}{\textbf{Phi}}     & \multicolumn{1}{c|}{\textbf{R1/R2/RL}}  & \multicolumn{1}{c|}{\begin{tabular}[c]{@{}c@{}}33.8/\\ 11.3\end{tabular}}                         & \multicolumn{1}{c|}{\begin{tabular}[c]{@{}c@{}}34.0/\\ 11.3\end{tabular}}                         & \multicolumn{1}{c|}{\begin{tabular}[c]{@{}c@{}}35.0/\\ 12.22\end{tabular}}                                 & \multicolumn{1}{c|}{\begin{tabular}[c]{@{}c@{}}36.5/\\ 12.61\end{tabular}}                        & \multicolumn{1}{c|}{\begin{tabular}[c]{@{}c@{}}34.8/\\ 11.82\end{tabular}}                                 & \multicolumn{1}{c|}{\begin{tabular}[c]{@{}c@{}}36.4/\\ 12.18\end{tabular}}                        & \multicolumn{1}{c|}{\textbf{\begin{tabular}[c]{@{}c@{}}37.5/\\ 13.75\end{tabular}}}               & \begin{tabular}[c]{@{}c@{}}35.5/\\ 13.23\end{tabular}                        \\ \cline{2-10} 
                                  & \multicolumn{1}{c|}{\textbf{METEOR}}    & \multicolumn{1}{c|}{26.62}                                                                        & \multicolumn{1}{c|}{22.35}                                                                        & \multicolumn{1}{c|}{26.89}                                                                                 & \multicolumn{1}{c|}{22.05}                                                                        & \multicolumn{1}{c|}{\textbf{27.8}}                                                                         & \multicolumn{1}{c|}{24.73}                                                                        & \multicolumn{1}{c|}{27.67}                                                                        & 21.75                                                                        \\ \cline{2-10} 
                                  & \multicolumn{1}{c|}{\textbf{BertScore}} & \multicolumn{1}{c|}{82.54}                                                                        & \multicolumn{1}{c|}{82.47}                                                                        & \multicolumn{1}{c|}{82.84}                                                                                 & \multicolumn{1}{c|}{\textbf{82.94}}                                                               & \multicolumn{1}{c|}{82.45}                                                                                 & \multicolumn{1}{c|}{82.44}                                                                        & \multicolumn{1}{c|}{82.81}                                                                        & 82.77                                                                        \\ \hline
\multirow{3}{*}{\textbf{Mistral}} & \multicolumn{1}{c|}{\textbf{R1/R2}}     & \multicolumn{1}{c|}{\begin{tabular}[c]{@{}c@{}}39.7/16.36\\ 39.8/19.81\\ 38.4/18.84\end{tabular}} & \multicolumn{1}{c|}{\begin{tabular}[c]{@{}c@{}}35.2/17.67\\ 35.2/18.62\\ 34.2/17.63\end{tabular}} & \multicolumn{1}{c|}{\textbf{\begin{tabular}[c]{@{}c@{}}42.5/17.53\\ 42.0/20.87\\ 42.3/17.66\end{tabular}}} & \multicolumn{1}{c|}{\begin{tabular}[c]{@{}c@{}}40.7/16.74\\ 37.5/19.49\\ 40.0/16.93\end{tabular}} & \multicolumn{1}{c|}{\begin{tabular}[c]{@{}c@{}}39.1/15.98\\ 39.5/19.22\\ 41.0/17.53\end{tabular}}          & \multicolumn{1}{c|}{\begin{tabular}[c]{@{}c@{}}42.1/18.62\\ 37.0/19.18\\ 42.0/18.57\end{tabular}} & \multicolumn{1}{c|}{\begin{tabular}[c]{@{}c@{}}41.2/16.53\\ 40.8/19.38\\ 41.0/16.64\end{tabular}} & \begin{tabular}[c]{@{}c@{}}40.0/16.61\\ 34.9/18.18\\ 39.8/16.89\end{tabular} \\ \cline{2-10} 
                                  & \multicolumn{1}{c|}{\textbf{METEOR}}    & \multicolumn{1}{c|}{\begin{tabular}[c]{@{}c@{}}28.35\\ 26.63\\ 26.58\end{tabular}}                & \multicolumn{1}{c|}{\begin{tabular}[c]{@{}c@{}}22.38\\ 21.46\\ 21.95\end{tabular}}                & \multicolumn{1}{c|}{\begin{tabular}[c]{@{}c@{}}29.0\\ 26.84\\ 28.8\end{tabular}}                           & \multicolumn{1}{c|}{\begin{tabular}[c]{@{}c@{}}23.51\\ 21.08\\ 23.27\end{tabular}}                & \multicolumn{1}{c|}{\textbf{\begin{tabular}[c]{@{}c@{}}29.74\\ 26.56\\ 29.72\end{tabular}}}                & \multicolumn{1}{c|}{\begin{tabular}[c]{@{}c@{}}26.35\\ 22.25\\ 26.35\end{tabular}}                & \multicolumn{1}{c|}{\begin{tabular}[c]{@{}c@{}}29.39\\ 26.44\\ 29.25\end{tabular}}                & \begin{tabular}[c]{@{}c@{}}25.02\\ 20.3\\ 24.75\end{tabular}                 \\ \cline{2-10} 
                                  & \multicolumn{1}{c|}{\textbf{BertScore}} & \multicolumn{1}{c|}{\begin{tabular}[c]{@{}c@{}}83.22\\ 83.24\\ 83.33\end{tabular}}                & \multicolumn{1}{c|}{\begin{tabular}[c]{@{}c@{}}83.31\\ 83.18\\ 83.29\end{tabular}}                & \multicolumn{1}{c|}{\begin{tabular}[c]{@{}c@{}}83.11\\ 83.16\\ 83.11\end{tabular}}                         & \multicolumn{1}{c|}{\begin{tabular}[c]{@{}c@{}}83.18\\ 83.23\\ 83.16\end{tabular}}                & \multicolumn{1}{c|}{\begin{tabular}[c]{@{}c@{}}83.57\\ 83.58\\ 83.67\end{tabular}}                         & \multicolumn{1}{c|}{\textbf{\begin{tabular}[c]{@{}c@{}}83.63\\ 83.52\\ 83.63\end{tabular}}}       & \multicolumn{1}{c|}{\begin{tabular}[c]{@{}c@{}}83.24\\ 83.13\\ 83.24\end{tabular}}                & \begin{tabular}[c]{@{}c@{}}83.28\\ 83.24\\ 83.27\end{tabular}                \\ \hline
\multirow{3}{*}{\textbf{o3}}      & \multicolumn{1}{c|}{\textbf{R1/R2}}     & \multicolumn{1}{c|}{\begin{tabular}[c]{@{}c@{}}37.99 ± 0.40\\ 12.54 ± 0.17\end{tabular}}          & \multicolumn{1}{c|}{\begin{tabular}[c]{@{}c@{}}38.10 ± 0.42\\ 11.90 ± 0.27\end{tabular}}          & \multicolumn{1}{c|}{\begin{tabular}[c]{@{}c@{}}37.79 ± 0.15\\ 12.14 ± 0.23\end{tabular}}                   & \multicolumn{1}{c|}{\begin{tabular}[c]{@{}c@{}}37.52 ± 0.38\\ 11.55 ± 0.12\end{tabular}}          & \multicolumn{1}{c|}{\begin{tabular}[c]{@{}c@{}}37.23 ± 0.36\\ 12.42 ± 0.09\end{tabular}}                   & \multicolumn{1}{c|}{\begin{tabular}[c]{@{}c@{}}38.63 ± 0.24\\ 12.56 ± 0.17\end{tabular}}          & \multicolumn{1}{c|}{\begin{tabular}[c]{@{}c@{}}37.97 ± 0.36\\ 12.62 ± 0.26\end{tabular}}          & \textbf{\begin{tabular}[c]{@{}c@{}}38.78 ± 0.23\\ 12.38 ± 0.15\end{tabular}} \\ \cline{2-10} 
                                  & \multicolumn{1}{c|}{\textbf{METEOR}}    & \multicolumn{1}{c|}{28.84 ± 0.25}                                                                 & \multicolumn{1}{c|}{25.43 ± 0.23}                                                                 & \multicolumn{1}{c|}{28.36 ± 0.16}                                                                          & \multicolumn{1}{c|}{23.72 ± 0.24}                                                                 & \multicolumn{1}{c|}{30.32 ± 0.15}                                                                          & \multicolumn{1}{c|}{28.34 ± 0.25}                                                                 & \multicolumn{1}{c|}{30.24 ± 0.26}                                                                 & 27.49 ± 0.13                                                                 \\ \cline{2-10} 
                                  & \multicolumn{1}{c|}{\textbf{BertScore}} & \multicolumn{1}{c|}{\textbf{82.95 ± 0.07}}                                                        & \multicolumn{1}{c|}{82.93 ± 0.04}                                                                 & \multicolumn{1}{c|}{82.90 ± 0.04}                                                                          & \multicolumn{1}{c|}{82.91 ± 0.08}                                                                 & \multicolumn{1}{c|}{82.94 ± 0.06}                                                                          & \multicolumn{1}{c|}{82.92 ± 0.08}                                                                 & \multicolumn{1}{c|}{82.81 ± 0.06}                                                                 & 82.70 ± 0.05                                                                 \\ \hline
\multicolumn{1}{|l|}{}            & \multicolumn{9}{c|}{\textbf{Dataset 2}}                                                                                                                                                                                                                                                                                                                                                                                                                                                                                                                                                                                                                                                                                                                                                                                                                              \\ \hline
\multirow{3}{*}{\textbf{Llama}}   & \multicolumn{1}{c|}{\textbf{R1/R2}}     & \multicolumn{1}{c|}{\begin{tabular}[c]{@{}c@{}}37.6/\\ 14.54\end{tabular}}                        & \multicolumn{1}{c|}{\begin{tabular}[c]{@{}c@{}}20.9/\\ 7.08\end{tabular}}                         & \multicolumn{1}{c|}{\begin{tabular}[c]{@{}c@{}}37.0/\\ 14.09\end{tabular}}                                 & \multicolumn{1}{c|}{\begin{tabular}[c]{@{}c@{}}16.5/\\ 5.61\end{tabular}}                         & \multicolumn{1}{c|}{\begin{tabular}[c]{@{}c@{}}38.2/\\ 13.76\end{tabular}}                                 & \multicolumn{1}{c|}{\begin{tabular}[c]{@{}c@{}}28.0/\\ 9.94\end{tabular}}                         & \multicolumn{1}{c|}{\textbf{\begin{tabular}[c]{@{}c@{}}40.8/\\ 14.66\end{tabular}}}               & \begin{tabular}[c]{@{}c@{}}31.2/\\ 10.05\end{tabular}                        \\ \cline{2-10} 
                                  & \multicolumn{1}{c|}{\textbf{METEOR}}    & \multicolumn{1}{c|}{21.44}                                                                        & \multicolumn{1}{c|}{10.85}                                                                        & \multicolumn{1}{c|}{19.67}                                                                                 & \multicolumn{1}{c|}{7.56}                                                                         & \multicolumn{1}{c|}{22.89}                                                                                 & \multicolumn{1}{c|}{15.47}                                                                        & \multicolumn{1}{c|}{\textbf{24.64}}                                                               & 17.08                                                                        \\ \cline{2-10} 
                                  & \multicolumn{1}{c|}{\textbf{BertScore}} & \multicolumn{1}{c|}{81.81}                                                                        & \multicolumn{1}{c|}{80.84}                                                                        & \multicolumn{1}{c|}{\textbf{82.11}}                                                                        & \multicolumn{1}{c|}{81.12}                                                                        & \multicolumn{1}{c|}{81.76}                                                                                 & \multicolumn{1}{c|}{81.57}                                                                        & \multicolumn{1}{c|}{81.5}                                                                         & 81.52                                                                        \\ \hline
\multirow{3}{*}{\textbf{Phi}}     & \multicolumn{1}{c|}{\textbf{R1/R2}}     & \multicolumn{1}{c|}{\begin{tabular}[c]{@{}c@{}}33.4/\\ 11.19\end{tabular}}                        & \multicolumn{1}{c|}{\begin{tabular}[c]{@{}c@{}}23.0/\\ 7.19\end{tabular}}                         & \multicolumn{1}{c|}{\begin{tabular}[c]{@{}c@{}}32.3/\\ 10.46\end{tabular}}                                 & \multicolumn{1}{c|}{\begin{tabular}[c]{@{}c@{}}19.7/\\ 6.21\end{tabular}}                         & \multicolumn{1}{c|}{\textbf{\begin{tabular}[c]{@{}c@{}}36.9/\\ 12.11\end{tabular}}}                        & \multicolumn{1}{c|}{\begin{tabular}[c]{@{}c@{}}30.4/\\ 10.37\end{tabular}}                        & \multicolumn{1}{c|}{\textbf{\begin{tabular}[c]{@{}c@{}}36.9/\\ 12.59\end{tabular}}}               & \begin{tabular}[c]{@{}c@{}}28.7/\\ 9.58\end{tabular}                         \\ \cline{2-10} 
                                  & \multicolumn{1}{c|}{\textbf{METEOR}}    & \multicolumn{1}{c|}{21.49}                                                                        & \multicolumn{1}{c|}{13.64}                                                                        & \multicolumn{1}{c|}{20.11}                                                                                 & \multicolumn{1}{c|}{11.12}                                                                        & \multicolumn{1}{c|}{22.37}                                                                                 & \multicolumn{1}{c|}{16.34}                                                                        & \multicolumn{1}{c|}{\textbf{22.57}}                                                               & 15.84                                                                        \\ \cline{2-10} 
                                  & \multicolumn{1}{c|}{\textbf{BertScore}} & \multicolumn{1}{c|}{80.88}                                                                        & \multicolumn{1}{c|}{80.5}                                                                         & \multicolumn{1}{c|}{80.25}                                                                                 & \multicolumn{1}{c|}{80.07}                                                                        & \multicolumn{1}{c|}{\textbf{82.1}}                                                                         & \multicolumn{1}{c|}{82.04}                                                                        & \multicolumn{1}{c|}{81.99}                                                                        & 81.89                                                                        \\ \hline
\multirow{3}{*}{\textbf{Mistral}} & \multicolumn{1}{c|}{\textbf{R1/R2}}     & \multicolumn{1}{c|}{\begin{tabular}[c]{@{}c@{}}38.3/13.64\\ 37.1/13.39\\ 37.4/13.37\end{tabular}} & \multicolumn{1}{c|}{\begin{tabular}[c]{@{}c@{}}14.3/4.39\\ 15.1/4.18\\ 14.2/4.41\end{tabular}}    & \multicolumn{1}{c|}{\begin{tabular}[c]{@{}c@{}}40.5/14.9\\ 39.0/14.77\\ 40.2/15.28\end{tabular}}           & \multicolumn{1}{c|}{\begin{tabular}[c]{@{}c@{}}17.0/5.91\\ 17.4/5.97\\ 17.2/5.97\end{tabular}}    & \multicolumn{1}{c|}{\textbf{\begin{tabular}[c]{@{}c@{}}42.4/16.33\\ 40.5/16.28\\ 42.4/16.31\end{tabular}}} & \multicolumn{1}{c|}{\begin{tabular}[c]{@{}c@{}}33.2/13.65\\ 32.1/13.58\\ 33.3/13.64\end{tabular}} & \multicolumn{1}{c|}{\begin{tabular}[c]{@{}c@{}}41.4/14.86\\ 39.7/14.27\\ 41.3/14.78\end{tabular}} & \begin{tabular}[c]{@{}c@{}}32.8/12.67\\ 31.2/11.51\\ 32.2/12.34\end{tabular} \\ \cline{2-10} 
                                  & \multicolumn{1}{c|}{\textbf{METEOR}}    & \multicolumn{1}{c|}{\begin{tabular}[c]{@{}c@{}}20.81\\ 20.45\\ 20.26\end{tabular}}                & \multicolumn{1}{c|}{\begin{tabular}[c]{@{}c@{}}6.7\\ 7.13\\ 6.7\end{tabular}}                     & \multicolumn{1}{c|}{\begin{tabular}[c]{@{}c@{}}22.06\\ 21.07\\ 21.67\end{tabular}}                         & \multicolumn{1}{c|}{\begin{tabular}[c]{@{}c@{}}7.47\\ 7.64\\ 7.59\end{tabular}}                   & \multicolumn{1}{c|}{\begin{tabular}[c]{@{}c@{}}25.78\\ 24.77\\ 25.78\end{tabular}}                         & \multicolumn{1}{c|}{\begin{tabular}[c]{@{}c@{}}17.15\\ 16.64\\ 17.15\end{tabular}}                & \multicolumn{1}{c|}{\begin{tabular}[c]{@{}c@{}}23.61\\ 22.36\\ 23.61\end{tabular}}                & \begin{tabular}[c]{@{}c@{}}15.89\\ 14.83\\ 15.74\end{tabular}                \\ \cline{2-10} 
                                  & \multicolumn{1}{c|}{\textbf{BertScore}} & \multicolumn{1}{c|}{\begin{tabular}[c]{@{}c@{}}82.28\\ 82.28\\ 82.11\end{tabular}}                & \multicolumn{1}{c|}{\begin{tabular}[c]{@{}c@{}}81.24\\ 81.28\\ 81.24\end{tabular}}                & \multicolumn{1}{c|}{\begin{tabular}[c]{@{}c@{}}82.25\\ 82.46\\ 82.52\end{tabular}}                         & \multicolumn{1}{c|}{\begin{tabular}[c]{@{}c@{}}81.6\\ 81.63\\ 81.59\end{tabular}}                 & \multicolumn{1}{c|}{\begin{tabular}[c]{@{}c@{}}82.98\\ 83.08\\ 82.98\end{tabular}}                         & \multicolumn{1}{c|}{\textbf{\begin{tabular}[c]{@{}c@{}}82.99\\ 83.06\\ 82.99\end{tabular}}}       & \multicolumn{1}{c|}{\begin{tabular}[c]{@{}c@{}}82.54\\ 82.38\\ 82.54\end{tabular}}                & \begin{tabular}[c]{@{}c@{}}82.47\\ 82.37\\ 82.52\end{tabular}                \\ \hline
\multirow{3}{*}{\textbf{o3}}      & \multicolumn{1}{c|}{\textbf{R1/R2}}     & \multicolumn{1}{c|}{\begin{tabular}[c]{@{}c@{}}36.56 ± 0.38\\ 11.60 ± 0.33\end{tabular}}          & \multicolumn{1}{c|}{\begin{tabular}[c]{@{}c@{}}26.04 ± 0.90\\ 7.83 ± 0.54\end{tabular}}           & \multicolumn{1}{c|}{\begin{tabular}[c]{@{}c@{}}36.90 ± 0.30\\ 12.04 ± 0.28\end{tabular}}                   & \multicolumn{1}{c|}{\begin{tabular}[c]{@{}c@{}}24.25 ± 0.89\\ 7.54 ± 0.28\end{tabular}}           & \multicolumn{1}{c|}{\begin{tabular}[c]{@{}c@{}}42.18 ± 0.30\\ 13.99 ± 0.14\end{tabular}}                   & \multicolumn{1}{c|}{\begin{tabular}[c]{@{}c@{}}39.64 ± 1.00\\ 13.15 ± 0.38\end{tabular}}          & \multicolumn{1}{c|}{\textbf{\begin{tabular}[c]{@{}c@{}}43.12 ± 0.26\\ 14.63 ± 0.30\end{tabular}}} & \begin{tabular}[c]{@{}c@{}}40.44 ± 0.55\\ 13.40 ± 0.34\end{tabular}          \\ \cline{2-10} 
                                  & \multicolumn{1}{c|}{\textbf{METEOR}}    & \multicolumn{1}{c|}{20.27 ± 0.40}                                                                 & \multicolumn{1}{c|}{12.65 ± 0.63}                                                                 & \multicolumn{1}{c|}{20.29 ± 0.15}                                                                          & \multicolumn{1}{c|}{11.18 ± 0.50}                                                                 & \multicolumn{1}{c|}{26.59 ± 0.39}                                                                          & \multicolumn{1}{c|}{22.59 ± 0.91}                                                                 & \multicolumn{1}{c|}{\textbf{27.34 ± 0.15}}                                                        & 22.75 ± 0.58                                                                 \\ \cline{2-10} 
                                  & \multicolumn{1}{c|}{\textbf{BertScore}} & \multicolumn{1}{c|}{\textbf{82.34 ± 0.09}}                                                        & \multicolumn{1}{c|}{81.96 ± 0.16}                                                                 & \multicolumn{1}{c|}{82.33 ± 0.07}                                                                          & \multicolumn{1}{c|}{81.84 ± 0.06}                                                                 & \multicolumn{1}{c|}{82.23 ± 0.05}                                                                          & \multicolumn{1}{c|}{82.33 ± 0.07}                                                                 & \multicolumn{1}{c|}{82.15 ± 0.12}                                                                 & 82.01 ± 0.11                                                                 \\ \hline
\end{tabular}
}
\caption{Results on judgment explanations for Llama, Phi, Mistral, and o3-mini are indicated, where R1/R2 refer to the Rouge-1 and Rouge-2 metrics, respectively. The parameters D, C, and R denote the presence of certain components in the experiments. In the case of the Mistral row, each cell contains three additional rows showing results for: (i) independent, where chains were evaluated with its corresponding binary outputs; (ii) common, involving unified test points across all chains that produced a binary result; and (iii) chainwise, which compares the performance of chained vs. non-chained counterparts. }
\label{both_datasets_expl}
\end{table}